\documentclass[preprint,12pt]{elsarticle}

\usepackage{adjustbox}
\usepackage{multirow}
\usepackage{pifont}
\usepackage{amssymb} % 或者 \usepackage{amsfonts}
\usepackage{amsmath}
\usepackage{cleveref}
\usepackage{CJKutf8}
\usepackage{xcolor}
\usepackage[ruled,linesnumbered]{algorithm2e}

\usepackage{algpseudocode} % For pseudocode commands
\usepackage{lineno}
\usepackage[normalem]{ulem}

\newcommand{\HL}[1]{#1}

\definecolor{r}{RGB}{255,0,0}
\definecolor{g}{RGB}{0,255,0}
\definecolor{b}{RGB}{0,0,255}
\usepackage{environ}
\newif\ifshowchanges
\showchangesfalse % 隐藏痕迹

\newcommand{\added}[1]{%
  \ifshowchanges
    \textcolor{blue}{#1}%
  \else
    #1%
  \fi
}

\newcommand{\deleted}[1]{%
  \ifshowchanges
    \textcolor{red}{\sout{#1}}%
  \fi
}

\newcommand{\replaced}[2]{%
  \ifshowchanges
    \deleted{#1} $\rightarrow$ \added{#2}%
  \else
    #2%
  \fi
}

\journal{Knowledge-Based Systems}

\begin{document}

\begin{frontmatter}

%% Title, authors and addresses

%% use the tnoteref command within \title for footnotes;
%% use the tnotetext command for theassociated footnote;
%% use the fnref command within \author or \affiliation for footnotes;
%% use the fntext command for theassociated footnote;
%% use the corref command within \author for corresponding author footnotes;
%% use the cortext command for theassociated footnote;
%% use the ead command for the email address,
%% and the form \ead[url] for the home page:
%% \title{Title\tnoteref{label1}}
%% \tnotetext[label1]{}
%% \author{Name\corref{cor1}\fnref{label2}}
%% \ead{email address}
%% \ead[url]{home page}
%% \fntext[label2]{}
%% \cortext[cor1]{}
%% \affiliation{organization={},
%%             addressline={},
%%             city={},
%%             postcode={},
%%             state={},
%%             country={}}
%% \fntext[label3]{}

\title{CCSD: Cross-Modal Compositional Self-Distillation for Robust Brain Tumor Segmentation with Missing Modalities} %% Article title

%% use optional labels to link authors explicitly to addresses:
%% \author[label1,label2]{}
%% \affiliation[label1]{organization={},
%%             addressline={},
%%             city={},
%%             postcode={},
%%             state={},
%%             country={}}
%%
%% \affiliation[label2]{organization={},
%%             addressline={},
%%             city={},
%%             postcode={},
%%             state={},
%%             country={}}

% \author{} %% Author name
\author[1]{Dongqing Xie}
\ead{2311844@tongji.edu.cn}

\author[3]{Yonghuang Wu}
\ead{24110720122@m.fudan.edu.cn}

\author[4]{Zisheng Ai}
\ead{azs1966@tongji.edu.cn}

\author[5]{Jun Min}
\ead{minjun@cjlu.edu.cn}

\author[1]{Zhencun Jiang}
\ead{zhencunjiang@tongji.edu.cn}

\author[2]{Shaojin Geng}
\ead{shaojin_geng@163.com}

\author[1,2]{Lei Wang\corref{cor1}}
\ead{wanglei@tongji.edu.cn}

\cortext[cor1]{Corresponding author.}

%% Author affiliations
\address[1]{%
Shanghai Research Institute for Intelligent Autonomous Systems, Tongji University,\\
Rui'an Building, 1239 Siping Road, Shanghai 200092, China
}

\address[2]{%
College of Electronic and Information Engineering, Tongji University,\\
Rui'an Building, 1239 Siping Road, Shanghai 201804, China
}

\address[3]{%
College of Biomedical Engineering, Fudan University,\\
2005 Songhu Road, Yangpu District, Shanghai 200438, China
}

\address[4]{%
School of Medicine, Tongji University,\\
Rui'an Building, 1239 Siping Road, Shanghai 200092, China
}

\address[5]{%
College of Information Engineering, China Jiliang University,\\
258 Xueyuan Street, Jianggan District, Hangzhou 310018, China
}

%% Abstract
\begin{abstract}
%% Text of abstract
The accurate segmentation of brain tumors from multi-modal MRI is critical for clinical diagnosis and treatment planning. While integrating complementary information from various MRI sequences is a common practice, the frequent absence of one or more modalities in real-world clinical settings poses a significant challenge, severely compromising the performance and generalizability of deep learning-based segmentation models.
To address this challenge, we propose a novel Cross-Modal Compositional Self-Distillation (CCSD) framework that can flexibly handle arbitrary combinations of input modalities. CCSD adopts a shared-specific encoder-decoder architecture and incorporates two self-distillation strategies: (i) a hierarchical modality self-distillation mechanism that transfers knowledge across modality hierarchies to reduce semantic discrepancies, and (ii) a progressive modality combination distillation approach that enhances robustness to missing modalities by simulating gradual modality dropout during training.
Extensive experiments on public brain tumor segmentation benchmarks demonstrate that CCSD achieves state-of-the-art performance across various missing-modality scenarios, with strong generalization and stability. % The framework shows great promise for practical deployment in clinically realistic imaging conditions.

\end{abstract}

%%Graphical abstract
% \begin{graphicalabstract}
% %\includegraphics{grabs}
% \end{graphicalabstract}

%%Research highlights
% \begin{highlights}
% \item Research highlight 1
% \item Research highlight 2
% \end{highlights}

\begin{highlights}
\item CCSD: a teacher-free cross-modal compositional self-distillation framework for multi-modal MRI brain tumor segmentation that supports arbitrary modality subsets.
\item Built upon an existing shared–specific encoder–decoder backbone to flexibly handle missing modalities while enabling efficient knowledge transfer.
\item Hierarchical Modality Self-Distillation transfers knowledge from full to partial modality sets, reducing semantic discrepancies across modality hierarchies.
\item Decremental Modality Combination Distillation simulates progressive modality loss via an optimized decrement path and sequential distillation, enhancing robustness.
\item State-of-the-art performance and stability across diverse missing-modality scenarios on public benchmarks, with strong generalization and clinical practicality (no external teacher or reconstruction pretraining).
\end{highlights}

%% Keywords
\begin{keyword}
%% keywords here, in the form: keyword \sep keyword
Multi-modal MRI \sep Missing modality \sep Cross-modal self-distillation

%% PACS codes here, in the form: \PACS code \sep code

%% MSC codes here, in the form: \MSC code \sep code
%% or \MSC[2008] code \sep code (2000 is the default)

\end{keyword}

\end{frontmatter}

\section{Introduction}
\label{sec-intro}

Deep learning has achieved remarkable success in medical image analysis, particularly in brain tumor segmentation tasks. Multi-modal MRI, due to its ability to provide complementary tissue contrast information for precise identification of tumor boundaries and sub-regions, has become the preferred imaging modality for clinical diagnosis \cite{bakas2018identifying, chen2022m3ae}. 
Advanced CNN and transformer architectures by significantly enhance multi-modal MRI segmentation performance through feature fusion and attention mechanisms \cite{chen2020brain, zhou2020one, LIN2024108591, tian2022axial}. 
\begin{figure}
    \centering
    \includegraphics[width=1\linewidth]{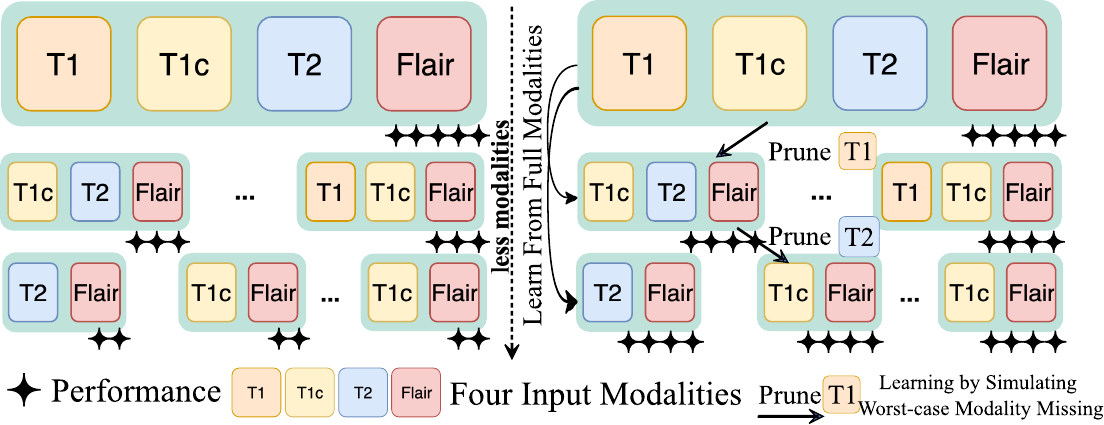}
    \caption{
      Left: Traditional methods handle modality missing scenarios but lack interaction mechanisms among different modality combinations, 
      resulting in significant performance drops when fewer modalities are available. Right: 
      \replaced{
        We propose a cross-modal combination self-distillation approach that introduces two strategies to enable knowledge transfer across hierarchical modality combinations, 
        and simulates modality missing during training to improve robustness and maintain performance under partial modality inputs.
      }{
        We propose a cross-modal combination self-distillation framework that (1) learns from full-modality supervision to transfer comprehensive multimodal knowledge, 
        and (2) improves robustness by explicitly simulating worst-case modality-missing scenarios during training, thereby maintaining strong performance under partial modality inputs.
      }
    }
    \label{fig:inno}
\end{figure}

However, in clinical practice, acquiring all four MRI modalities (FLAIR, T1, T1c, T2) is often impossible due to motion, artifacts, or equipment issues. This common modality missing problem severely degrades deep learning performance \cite{azad2025addressing, wu2024comprehensive} and limits clinical reliability, as current fusion methods assume full modality availability during training and inference.

Handling missing modalities in multimodal learning remains a persistent challenge, primarily due to the insufficient flexibility of existing methods in adapting to partial inputs—such as \cite{chen2022m3ae}, which relies on image reconstruction pretraining tasks to accommodate modality absence, or the inadequate cross-modal knowledge transfer, as seen in \cite{wang2023multi}, which investigates the lightweight disentanglement and fusion of shared and specific features 
yet the potential of inter-modal knowledge transfer in enhancing robustness under realistic modality dropout scenarios remains underexplored.
Although these issues are widely acknowledged, they have not been adequately addressed by current approaches.

To address these challenges, we propose a novel Cross-Modal Compositional Self-Distillation (CCSD) framework that enhances segmentation performance and robustness under arbitrary modality missing scenarios via a flexible and efficient knowledge transfer mechanism (see Figure~\ref{fig:inno}). Our approach utilizes a Shared-Specific encoder-decoder architecture that effectively disentangles and fuses multi-modal features. 
Built upon this, we introduce two innovative self-distillation strategies to enhance the model's performance and robustness in handling multi-modal data: Hierarchical Modality Self-Distillation (HMSD) and Decremental Modality Combination Distillation (DMCD). 
HMSD bridges semantic gaps across modality hierarchies by transferring knowledge from complete to partial modality sets, enabling richer representation learning. DMCD simulates realistic gradual modality failure by constructing an optimal decrement path, through which sequential distillation progressively strengthens model robustness to partial data availability.

The main contributions of this paper are summarized as follows:
\begin{itemize}
    \item We proposed a flexible framework to handle arbitrary modality combinations via shared-specific encoders, enabling efficient self-distillation under missing modalities.
    
    \item We design a hierarchical modality self-distillation to transfer knowledge from full to partial modalities, reducing conflicts and improving feature consistency.
    
    \item We introduce a decremental modality combination distillation strategy to simulate realistic gradual modality failure and employ sequential distillation with dynamic path optimization to simulate progressive modality loss, thereby enhancing robustness.
    
\end{itemize}

\section{Related Work}
\label{sec-rw}

\subsection{Multi-modal Learning}

Modern brain tumor segmentation leverages multi-modal MRI (T1, T1c, T2, FLAIR) to capture complementary tumor characteristics. Early CNN-based methods employed dedicated fusion modules for cross-modal integration~\cite{zhang2021cross, zhou2021feature, ding2021rfnet}, later enhanced by attention mechanisms that adaptively weight modalities~\cite{LIN2024108591, tian2022axial}. Recently, Transformer-based~\cite{peiris2022hybrid, xing2022nestedformer, karimijafarbigloo2024mmcformer} and hybrid CNN-Transformer architectures~\cite{wang2021transbts, jia2021bitr} have improved long-range modeling and modality interaction. 
However, these approaches assume complete modality availability, a condition rarely satisfied in clinical practice due to motion artifacts, scanner variability, or protocol mismatches, resulting in significant performance degradation when any modality is absent~\cite{ding2021rfnet, karimijafarbigloo2024mmcformer}.

\subsection{Missing Modalities}

To handle incomplete inputs, existing methods fall into three categories. \textit{Modality augmentation} synthesizes missing images using GANs~\cite{sharma2019missing,yu20183d} or VAEs~\cite{hamghalam2021modality}, but often introduces artifacts and performs poorly with single-input scenarios. \textit{Feature space engineering} models inter-modal correlations in latent space through correlation constraints~\cite{ma2021maximum, liu2021incomplete}, lightweight fusion~\cite{zhang2024tmformer}, or feature generation~\cite{kim2024missing}. While avoiding explicit synthesis, these methods typically assume consistent missing patterns during training and testing; this limits generalization under rare or unseen combinations (e.g., T1-only). \textit{Architecture engineering} designs flexible models via masking mechanisms~\cite{qian2023contrastive}, shared representations~\cite{yao2024drfuse}, or multimodal Transformers~\cite{mordacq2024adapt} that support arbitrary inputs. Though robust in design, many require complex structures or predefined patterns, reducing scalability and impeding knowledge sharing across configurations. Alternative model selection strategies~\cite{wang2021acn, hu2020knowledge, li2023makes} further trade flexibility for specialization.

Overall, while progress has been made, most methods fail to exploit the shared semantics across different modality subsets, especially under diverse or unobserved missing patterns.

\subsection{\added{Knowledge Distillation}}

\added{Knowledge Distillation (KD) aims to transfer supervisory information from a powerful teacher model to weaker student models~\cite{hinton2015distilling}. In missing-modality medical image segmentation, KD is commonly used to propagate rich multi-modal knowledge from full-modality models to models operating on partial inputs~\cite{hu2020knowledge,wang2021imagine,chen2021learning,azad2022smu}. Most existing methods follow a \emph{teacher-driven} paradigm, where full-modality networks are trained independently and then used as fixed teachers for specific missing-modality students. While effective, this paradigm suffers from three key limitations: (i) it requires training and maintaining separate teacher models, increasing computational cost; (ii) knowledge transfer is confined to predefined teacher--student pairs, limiting interaction across modality subsets; and (iii) it does not scale well to arbitrary or unseen missing-modality combinations.}

\added{To reduce reliance on external teachers, self-distillation frameworks~\cite{chen2022m3ae} have been introduced, where different modality subsets share parameters within a single network and supervise each other implicitly. Although this strategy simplifies training, existing self-distillation methods still regard each modality combination as an independent learning target, lacking an explicit mechanism to align representations across modality cardinalities or to account for structured modality loss.}

\added{These observations motivate a shift from \emph{pairwise} or \emph{independent} distillation toward a \emph{compositional} perspective. Instead of treating modality subsets in isolation, we propose to organize them according to their inherent structural relationships and enable systematic knowledge exchange within a unified self-distillation framework. This design naturally supports hierarchical distillation from complete to partial modality sets and decremental distillation along progressively degraded modality paths, thereby facilitating efficient knowledge sharing, improved robustness to worst-case and unseen missing patterns, and zero additional cost at inference.}

\section{Method}
\label{sec-method}

\subsection{Overall Architecture}

\begin{figure*}
    \centering
    \includegraphics[width=1\linewidth]{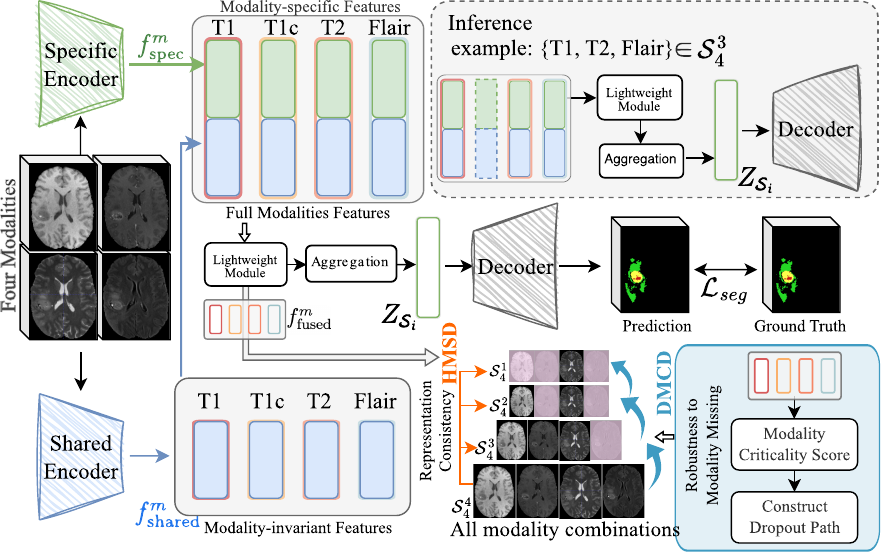}
    \caption{
      Overview. Left: We employs two encoders to capture both shared and specific features for each modality. 
    For all possible modality combinations, these two types of features are concatenated along both feature and channel dimensions, 
    followed by a lightweight convolutional layer to obtain the fused features for the corresponding modality combination. 
    Based on the fused features, we implement Hierarchical Modality Self-Distillation and Decremental Modality Combination Distillation. 
    The two distillation losses, together with the segmentation loss computed from the decoder using the fused features, 
    are jointly used to optimize the model parameters. During inference, the model can adapt to various scenarios involving missing modalities.   
    \deleted{
    Right: Algorithmic illustration of DMCD. $\mathcal{S}_N^n$: under ideal conditions with $N$ modalities, 
    the set of all cases containing only $n$ modalities. $\mathcal{S}_i$: a specific modality combination, $\tau$: temperature.}}
    \label{fig:method}
\end{figure*}

As shown in Figure~\ref{fig:method}, we propose a modality-aware encoder-decoder framework built upon a shared-specific disentanglement principle to systematically separate and fuse multi-modal features. The architecture operates as follows: Modality-specific encoders $E_{\text{spec}}^m$ (one per modality $m \in \{\text{FLAIR, T1, T1c, T2}\}$) first extract domain-adaptive features $f_{\text{spec}}^m = E_{\text{spec}}^m(x^m)$ that encode unique semantic patterns inherent to each imaging sequence. Concurrently, all modalities are processed through a single shared encoder $E_{\text{shared}}$ to derive invariant low-level representations \(f_{\text{shared}}^m = E_{\text{shared}}(x^m)\), establishing a common feature foundation. 
Crucially, a learnable compositional layer $C(\cdot)$ then fuses these complementary representations into a discriminative hybrid feature $f_{\text{fused}}^m = C(f_{\text{shared}}^m, f_{\text{spec}}^m)$, explicitly modeling both inter-modality commonality and intra-modality specificity.
\added{Specifically, the shared feature $f_{\text{shared}}^{m}$ and the modality-specific feature $f_{\text{spec}}^{m}$ are concatenated along the channel dimension and fed into a small projection (implemented by a $3\times3\times3$ 3D convolution plus normalization and non-linearity).}
\added{To support arbitrary modality combinations, we further aggregate modality-wise representations across all modalities to form a unified multi-modal feature $Z_{\mathcal{S}_i}$ corresponding to the modality set $\mathcal{S}_i$.}
\replaced{Finally, a shared decoder $D_{\text{shared}}$ reconstructs the target output from the fused features. This unified design leverages shared representations to ensure cross-modality consistency and modality-specific encoders to preserve specific characteristics, enabling the complementary shared and specific features to further facilitate flexible cross-modality self-distillation as formalized in following sections.}
{Finally, the unified representation $Z_{\mathcal{S}_i}$ is fed into a shared decoder $D_{\text{shared}}$ to reconstruct the target output. }

\subsection{Handling Full and Missing Modalities}

To enable the model to handle input with any combination of modalities, we design a unified forward propagation mechanism. Given a set of $N$ single modalities $\mathcal{M} = \{m_1, m_2, ..., m_N\}$, the set of all possible non-empty modality combinations is $\mathcal{S}$ ($|\mathcal{S}| = \sum_{k=1}^N \binom{N}{k} = 2^N - 1$). For any input modality subset $\mathcal{S}_i \subseteq \mathcal{S}$, our processing pipeline is as follows. 

\replaced{First, for any missing modality $m_j \notin \mathcal{S}_i$, we perform a masking operation by setting its corresponding input channel to zero ($x^{m_j} = 0$). Subsequently, the inputs of all modalities (whether present or missing) are passed through their respective encoders.}
{First, for any missing modality $m_j \notin \mathcal{S}_i$, we perform a masking operation by setting its corresponding input channel to zero ($x^{m_j} = 0$). Subsequently, the inputs of all modalities (whether present or missing) are passed through a shared encoder with identical parameters for all modalities.}

During the feature fusion stage, for present modalities $m_j \in \mathcal{S}_i$, we compute their fused feature $f_{fused}^{m_j}$. For the masked missing modalities $m_j \notin \mathcal{S}_i$, we directly use their shared feature $f_{shared}^{m_j}$ as their representation. Finally, we concatenate the features of all modalities to form a unified, high-dimensional feature vector $Z_{\mathcal{S}_i}$ that represents the current modality combination $\mathcal{S}_i$, which is then fed into the decoder.

It is noteworthy that in a single forward pass, we compute and cache the fused features corresponding to all $2^N-1$ possible non-empty modality combinations, storing them in a dictionary. This efficient implementation allows us to seamlessly and cost-effectively access the feature representations of any two modality combinations during the training process, facilitating the subsequent KD strategies (see Figure~\ref{fig:method}).

\paragraph{\added{Robustness-Oriented Training Protocol.}}\label{para:robustness-protocol}
To explicitly support arbitrary clinical input configurations, CCSD caches fused features for all non-empty modality combinations in each forward pass and performs distillation across hierarchies and decremental paths within the same iteration. During inference, the same unified fusion\,/\,decoder pipeline is applied to whichever modalities are available, without architectural changes or retraining.

\subsection{\added{Hierarchical Modality Self-Distillation}}

\added{Traditional multi-modal distillation methods typically perform knowledge transfer directly from the full modality set to a single modality or a fixed subset of modalities. Such direct distillation overlooks the inherent hierarchical relationships among different modality subsets and often leads to an abrupt semantic gap between the teacher and student models, especially when the student observes only limited modalities.}

\added{To address this issue, we propose \textbf{Hierarchical Modality Self-Distillation (HMSD)}, which introduces intermediate modality hierarchies to enable smooth and progressive knowledge transfer. In HMSD, the model that utilizes all available modalities consistently serves as the \textit{sole teacher}, while models with partial modality observations act exclusively as \textit{students}. Importantly, partial-modality models do not distill knowledge back to the teacher or among themselves, thereby preventing noise accumulation from incomplete modality inputs.}

\added{Specifically, under the default training condition where all $N$ modalities are available, we randomly sample a hierarchy level $k$ ($1 \leq k \leq N-1$) and construct all modality combinations consisting of exactly $k$ modalities:}
\begin{equation}
\mathcal{S}_N^k = \{ \mathcal{S}_i \mid |\mathcal{S}_i| = k \}.
\end{equation}
\added{For example, when $N=4$ and $k=3$, the resulting hierarchy includes $(1,2,3)$, $(1,2,4)$, $(1,3,4)$, and $(2,3,4)$.}

\replaced{During training, the full-modality model produces a soft output distribution and acts as a powerful teacher. Each student model corresponding to a modality subset $\mathcal{S}_i \in \mathcal{S}_N^k$ is trained in parallel to match the teacher's output through knowledge distillation. By allowing modality subsets of varying completeness to directly learn from the full-modality teacher, HMSD effectively bridges the semantic gap between complete and incomplete modality settings and enforces representation consistency across all modality subsets.}{During training, the full-modality configuration within the unified network produces soft output distributions and serves as the teacher; this teacher is not a separately trained model. Each student corresponding to a modality subset $\mathcal{S}_i \in \mathcal{S}_N^k$ is trained in parallel to match these teacher outputs via knowledge distillation. By allowing modality subsets of varying completeness to directly learn from the full-modality outputs produced by the same network, HMSD effectively bridges the semantic gap between complete and incomplete modality settings and enforces representation consistency across all modality subsets.}

\added{The hierarchical modality self-distillation loss is defined as:}
\begin{equation}
\mathcal{L}_{\text{HMSD}} = \frac{1}{|\mathcal{S}_N^k|} \sum_{\mathcal{S}_i \in \mathcal{S}_N^k}
D_{\text{KL}} \left( P_T \parallel P_{\mathcal{S}_i} \right),
\end{equation}
\added{where the two distributions are produced by the \emph{same unified network} under different modality configurations:}
\begin{equation}
P_T = \mathrm{Softmax}\!\left(D_{\text{shared}}(Z_{T})/\tau\right), \quad
P_{\mathcal{S}_i} = \mathrm{Softmax}\!\left(D_{\text{shared}}(Z_{\mathcal{S}_i})/\tau\right).
\end{equation}
\added{Here, $Z_{T}$ and $Z_{\mathcal{S}_i}$ are the fused representations under full and partial modality inputs, respectively, $D_{\text{shared}}(\cdot)$ is the shared decoder, and $\tau$ is the temperature parameter. The teacher $P_T$ corresponds to the full-modality configuration with stop-gradient applied, while partial-modality configurations are trained exclusively as students. No distillation signals are propagated among students or back to the teacher, ensuring stable hierarchical transfer without noise amplification.}

\added{Notably, the use of soft probability distributions naturally attenuates the impact of uncertain or noisy predictions from partial-modality observations, 
as such cases yield smoother distributions and smaller distillation gradients. Furthermore, averaging the distillation loss over all modality subsets 
at a given hierarchy level serves as an implicit regularization, reducing variance across different modality combinations. 
As a result, HMSD ensures stable and effective knowledge transfer while leveraging partial-modality supervision without introducing significant noise.}

\subsection{Decremental Modality Combination Distillation}
\label{sec:DMCD}

Although HMSD improves representation quality across multimodal combinations, it treats each missing combination as an independent scenario, failing to exploit structural relationships between related combinations (e.g., $\{T1,T2,FLAIR\}$ vs $\{T1,T2\}$) and underrepresenting high-frequency missing patterns. To bridge this gap, we propose \textbf{Decremental Modality Combination Distillation}, a training paradigm that deliberately simulates catastrophic data loss scenarios. As illustrated in Algorithm of the Figure~\ref{fig:method}, DMCD progressively removes the most task-critical modality at each step, starting from the full set $\mathcal{S}_i$, until only one modality remains. This systematic degradation forces the model to learn robust compensation strategies for irreversible information loss.

First, a \textbf{criticality score} $s(m_j)$ is defined for each modality $m_j \in \mathcal{S}_i$ to quantify its contribution to current task performance. A \textbf{higher score indicates greater task-criticality} (i.e., higher unique contribution and lower replaceability):
\begin{equation}
s(m_j) = - \sum_{z \neq j} \hat{s}(m_j, m_z)
\end{equation}
where $\hat{s}(m_j, m_z)$ is the feature cosine similarity between modalities $m_j$ and $m_z$:
\begin{equation}
\deleted{\hat{s}(m_j, m_z) = \frac{\langle F(m_j), F(m_z) \rangle}{\|F(m_j)\| \cdot \|F(m_z)\|}}
\added{\hat{s}(m_j, m_z) = \frac{\langle f_{fused}^{m_j}, f_{fused}^{m_z} \rangle}{\|f_{fused}^{m_j}\| \cdot \|f_{fused}^{m_z}\|}}
\end{equation}
\deleted{Here, $F(m_j)$ is the feature representation of modality $m_j$. }

At each step, the modality with the highest criticality score is removed to simulate worst-case data loss:
\begin{equation}
m_{\text{remove}} = \arg \max_{m_j \in \mathcal{S}_{i}} s(m_j) % \quad \text{\textit{(deliberate removal of most critical modality)}}
\end{equation}
The modality set is updated $\mathcal{S}_{i-1} = \mathcal{S}_{i} \setminus \{m_{\text{remove}}\}$. This generates a criticality-ordered path $P$ where each step represents a progressively more severe data failure scenario:
\begin{equation}
P = \{\mathcal{S}_i, \mathcal{S}_{i-1}, \mathcal{S}_{i-2}, ..., \mathcal{S}_1\}
\end{equation}

After constructing the path $P$, we perform sequential knowledge distillation in reverse order of severity. For adjacent combinations $(\mathcal{S}_k, \mathcal{S}_{k-1})$ where $\mathcal{S}_{k-1}$ is missing the most critical modality of $\mathcal{S}_k$. We treat the combination $\mathcal{S}_k$ as the teacher and $\mathcal{S}_{k-1}$ as the student. The distillation process encourages the student model's feature representation $Z_{\mathcal{S}_{k-1}}$ to mimic its teacher model's feature representation $Z_{\mathcal{S}_{k}}$ as closely as possible. The distillation process trains the student to reconstruct the teacher's representation using only residual modalities, effectively learning to compensate for the loss of irreplaceable information:
\begin{equation}
\mathcal{L}_{\text{DMCD}} = \sum_{k=2}^{\lvert P \rvert } D_{\text{KL}}\left( \sigma(Z_{\mathcal{S}_{k}} / \tau) \ \| \ \sigma(Z_{\mathcal{S}_{k-1}} / \tau) \right)
\end{equation}

\textbf{Why remove critical modalities?} Conventional approaches avoid removing high-contribution modalities as they are deemed "too important to lose". 
However, DMCD intentionally induces these worst-case scenarios during training, forcing the model to identify which modalities carry irreplaceable information and learn to reconstruct missing critical information from residual modalities.
This directly addresses the limitation of HMSD which only handles random missing patterns, not targeted loss of critical data (see the experimental section below).

\begin{algorithm}[H]
\caption{\added{Training Procedure of CCSD with HMSD and DMCD}}
\label{alg:ccsd}
\KwIn{Multi-modal input $\{x^m\}_{m \in \mathcal{M}}$, ground-truth label $y$}
\KwOut{Optimized network parameters $\theta$}

\For{each training iteration}{
    \tcp{Step 1: Forward pass and feature caching}
    \For{each modality $m \in \mathcal{M}$}{
        $f_{\text{shared}}^m \leftarrow E_{\text{shared}}(x^m)$\;
        $f_{\text{spec}}^m \leftarrow E_{\text{spec}}^m(x^m)$\;
        $f_{\text{fused}}^m \leftarrow C(f_{\text{shared}}^m, f_{\text{spec}}^m)$\;
    }
    % \textcolor{red}{Construct feature dictionary} $\mathcal{D} = \{Z_{\mathcal{S}} \mid \mathcal{S} \subseteq \mathcal{M}, \mathcal{S} \neq \emptyset\}$ by concatenating features of all modality subsets\;
    Construct feature dictionary $\mathcal{D} = \{Z_{\mathcal{S}_i} \mid \mathcal{S}_i \subseteq \mathcal{S}\}$ by concatenating features of all modality subsets\;

    \tcp{Step 2: Segmentation loss}
    Sample a modality subset $\mathcal{S}_i$\;
    $\hat{y}_{\mathcal{S}_i} \leftarrow D_{\text{shared}}(Z_{\mathcal{S}_i})$\;
    $\mathcal{L}_{\text{seg}} \leftarrow \ell(\hat{y}_{\mathcal{S}_i}, y)$\;

    \tcp{Step 3: Hierarchical Modality Self-Distillation (HMSD)}
    Sample hierarchy level $k \in \{1, \dots, N-1\}$\;
    $\mathcal{L}_{\text{HMSD}} \leftarrow \frac{1}{|\mathcal{S}_N^k|}
    \sum_{\mathcal{S}_j \in \mathcal{S}_N^k}
    D_{\text{KL}}\!\left(P_T \parallel P_{\mathcal{S}_j}\right)$\;
    \tcp{Teacher and student predictions are generated by the same network with stop-gradient on teacher}

    \tcp{Step 4: Decremental Modality Combination Distillation (DMCD)}
    % \textcolor{red}{Initialize} $\mathcal{S}_N = \mathcal{M}$\;
    Initialize $\mathcal{S}_i$\;

    Compute modality criticality scores $\{s(m)\}_{m \in \mathcal{S}_i}$ using feature cosine similarity\;
    Construct criticality-ordered path $P = \{\mathcal{S}_i, \dots, \mathcal{S}_1\}$ by iteratively removing the most critical modality\;
    $\mathcal{L}_{\text{DMCD}} \leftarrow \sum_{k=2}^{\lvert  P \rvert}
    D_{\text{KL}}\!\left(
    \sigma(Z_{\mathcal{S}_k}/\tau) \parallel \sigma(Z_{\mathcal{S}_{k-1}}/\tau)
    \right)$\;

    \tcp{Step 5: Joint optimization}
    $\mathcal{L}_{\text{total}} \leftarrow
    \mathcal{L}_{\text{seg}} + \mathcal{L}_{\text{HMSD}} + \mathcal{L}_{\text{DMCD}}$\;
    Update $\theta$ by backpropagating $\mathcal{L}_{\text{total}}$\;
    % lambda_1
}
\end{algorithm}

\section{Experiments}

\textbf{Datasets:} To enable a fair and reproducible comparison with existing methods, this study evaluates the proposed approach on two widely used public benchmarks: BraTS 2018 and BraTS 2020. The datasets contain 285 and 369 training cases, respectively, all of which are publicly available with expert-annotated ground truth labels. Following the consistent data partitioning protocol \cite{chen2022m3ae}, we split the BraTS 2018 dataset into 199 training, 29 validation, and 57 test cases, and the BraTS 2020 dataset into 219 training, 50 validation, and 100 test cases. The input consists of four MRI modalities have undergone standardized preprocessing including skull-stripping, spatial alignment, and isotropic resampling to $1~\text{mm}^3$ resolution. Evaluation is performed on three clinically significant tumor subregions: the whole tumor (WT), tumor core (TC), and enhancing tumor (ET), with performance quantified using the Dice Similarity Coefficient (Dice). 
Final results are reported on the held-out test set. This experimental setup aligns with recent state-of-the-art methods, ensuring a rigorous and objective evaluation under a unified benchmark.

\textbf{Implementation:} We adopt the same backbone network as ShaSpec~\cite{wang2023multi}, which is based on a 3D U-Net architecture. The encoder consists of a shared feature encoder and modality-specific feature encoders, with feature fusion taking place at the early stages of the decoder. The model is optimized using the Adam optimizer with an initial learning rate of $10^{-2}$, which is decayed using a cosine annealing schedule. Training is conducted for 100 epochs with a batch size of 12. Data augmentation, including random rotations and flips, is applied to enhance generalization. 
At inference, the model is evaluated under various missing modality scenarios. Experiments are performed on an NVIDIA A100 GPU. The segmentation performance is evaluated using the Dice score.

\subsection{Results}

Table~\ref{tab:2018} and Table~\ref{tab:2020} compare the performance of our framework on the BraTS 2018 and BraTS 2020 datasets, respectively. For BraTS 2018, we reference \cite{wang2023multi} and compare against the performance of four recent state-of-the-art methods for brain tumor segmentation with missing modalities: mmFm \cite{zhang2022mmformer}, ShaSpec \cite{wang2023multi}, M3AE \cite{chen2022m3ae}, and MIFPN \cite{diao2025multimodal}.
For BraTS 2020, we reference the baseline from \cite{chen2022m3ae} and compare against the following four methods: SMU-Net \cite{azad2022smu}, ShaSpec \cite{wang2023multi}, M3AE \cite{chen2022m3ae}, MIFPN \cite{diao2025multimodal} and \added{TMFormer \cite{zhang2024tmformer}.} 

\begin{table*}[]
\caption{Results on BraTS 2018. Modalities 1-4 denote FLAIR, T1, T1c, and T2, respectively. 
Results for M3AE$^\dagger$ is taken from \cite{chen2022m3ae}.
Metric: Dice score.}
\label{tab:2018}
\setlength{\tabcolsep}{3pt}
\begin{adjustbox}{max width=\textwidth}
\begin{tabular}{c|ccccc|ccccc|ccccc}
\hline
\multirow{2}{*}{\textbf{Modalities}} & \multicolumn{5}{c|}{\textbf{Enhancing Tumor}}        & \multicolumn{5}{c|}{\textbf{Tumor Core}}             & \multicolumn{5}{c}{\textbf{Whole Tumor}}           \\ \cline{2-16} 
                                     & mmFm    &ShaSpec    &M3AE$^\dagger$& MIFPN & \textbf{Ours}          & mmFm    & ShaSpec    &M3AE$^\dagger$& MIFPN &\textbf{Ours}          & mmFm   &ShaSpec    & M3AE$^\dagger$& MIFPN &\textbf{Ours}\\ \hline
1                                    & 39.33   & 37.79     & 35.60  & 42.28 & \textbf{45.48}         & 61.21   & 69.26      & 66.40 & 62.64 & \textbf{70.89}         & 86.10   & 89.65      & 88.70 & 86.79 & \textbf{90.40}   \\
2                                    & 32.53   & 38.43     & 37.10  & 40.46 & \textbf{42.82}         & 56.55   & \textbf{66.49}      & 66.10 & 57.01 & 65.20         & 67.52   & 73.66      & \textbf{74.40} & 67.75 & 74.17   \\
3                                    & 72.60   & \textbf{74.27}     & 73.70  & 72.87 & 74.22         & 75.41   & 81.85      & \textbf{82.90} & 78,86 & 78.05         & 72.22   & 74.35      & \textbf{75.80} & 73.09 & 75.52   \\
4                                    & 43.05   & \textbf{47.62}     & 47.60  & 45.49 & 46.76         & 64.20   & \textbf{70.90}      & 69.40 & 66.69 & 69.43         & 81.15   & 84.15      & \textbf{84.80} & 81.63 & 84.54   \\
12                                   & 42.96   & 43.85     & 41.20  & 44.55 & \textbf{45.58}         & 65.91   & 72.76      & 70.80 & 67.23 & \textbf{73.60}         & 87.06   & 88.36      & 89.00 & \textbf{89.85} & 88.93   \\
13                                   & 75.07   & 76.18     & 75.00  & 74.40 & \textbf{76.84}         & 77.88   & 84.11      & \textbf{84.40} & 78.13 & 84.31         & 87.30   & 90.33      & 89.70 & 89.72 & \textbf{90.46}   \\
14                                   & 47.52   & 44.66     & 45.40  & 47.09 & \textbf{48.81}         & 69.75   & 71.72      & 70.90 & 70.99 & \textbf{73.17}         & 87.59   & 90.14      & 89.90 & 89.49 & \textbf{92.04}   \\
23                                   & 74.04   & 74.53     & 74.70  & 74.28 & \textbf{77.01}         & 78.59   & 82.87      & 83.40 & 79.89 & \textbf{84.02}         & 74.42   & 77.51      & 77.20 & 77.03 & \textbf{78.63}   \\
24                                   & 44.99   & \textbf{48.90}     & 48.70  & 45.68 & 47.39         & 69.42   & 72.05      & 71.80 & \textbf{72.34} & 72.15         & 82.20   & \textbf{87.31}      & 86.70 & 84.61 & 86.47   \\
34                                   & 74.51   & 76.52     & 75.30  & 75.65 & \textbf{76.88}         & 78.61   & 83.44      & 84.20 & 81.16 & \textbf{84.79}         & 82.99   & 85.59      & \textbf{86.30} & 83.55 & 86.23   \\
123                                  & 75.47   & 75.87     & 74.00  & 76.41 & \textbf{77.63}         & 79.80   & 84.08      & 84.10 & 83.35 & \textbf{85.28}         & 87.33   & 89.02      & 88.90 & \textbf{90.01} & 89.20   \\
124                                  & \textbf{47.70}   & 44.66     & 44.80  & 45.33 & 47.30         & 71.52   & 73.38      & 72.70 & 74.41 & \textbf{74.64}         & 87.75   & 89.46      & 89.90 & 89.37 & \textbf{90.24}   \\
134                                  & 75.67   & 74.83     & 73.80  & 74.93 & \textbf{76.73}         & 79.55   & 85.69      & 84.60 & 83.03 & \textbf{86.86}         & 88.14   & 89.76      & 90.20 & 90.56 & \textbf{90.91}   \\
234                                  & 74.75   & 75.74     & 75.40  & 75.28 & \textbf{77.14}         & 80.39   & 84.79      & 84.40 & 83.24 & \textbf{85.42}         & 82.71   & 85.30      & 85.70 & 85.40 & \textbf{87.82}   \\
1234                                 & 77.61   & 76.47     & 75.50  & 76.92 & \textbf{79.95}         & 85.78   & 85.58      & 84.50 & 84.82 & \textbf{85.70}         & 89.64   & 90.15      & 90.10 & 90.59 & \textbf{91.50}   \\ \hline
Avg.                                 & 59.85   & 60.69     & 59.85  & 60.77 & \textbf{62.70}         & 72.97   & 77.93      & 77.37 & 74.63 & \textbf{78.23}         & 82.94   & 85.65      & 85.82 & 84.63 & \textbf{86.47}   \\ \hline
\end{tabular}
\end{adjustbox}
\end{table*}

\begin{table}[]
\caption{Results on BraTS 2020. Results for SMU-Net$^\dagger$ and M3AE$^\dagger$ are adopted from \cite{chen2022m3ae}. Metric: Dice $\uparrow$ and HD95 $\downarrow$.}
\label{tab:2020}
\centering
\begin{adjustbox}{max width=\columnwidth}
\begin{tabular}{c c|cccccc}
\hline
Method & Metric & SMU-Net & ShaSpec & M3AE & MIFPN & \added{TMFormer} & Ours \\ \hline
WT  & Dice  & 85.30$^\dagger$ & 86.27 & 86.90$^\dagger$ & 85.48 & 86.48    & \textbf{87.21} \\
TC  & Dice  & 77.70$^\dagger$ & 78.53 & 79.10$^\dagger$ & 78.00 & 80.92    & \textbf{82.54} \\
ET  & Dice  & 59.70$^\dagger$ & 59.91 & 61.70$^\dagger$ & 60.22 & 62.74    & \textbf{65.93} \\ \cline{1-8}
Avg.& Dice  & 74.23$^\dagger$ & 74.90 & 75.90$^\dagger$ & 74.57 & 76.68    & \textbf{78.56} \\ \hline
WT  & \HL{HD95}  & 7.30            & 6.50  & 6.90$^\dagger$  & 7.50  & 6.40     & \textbf{6.40} \\
TC  & \HL{HD95}  & 8.10            & 8.20  & 8.40$^\dagger$  & 8.70  & 7.80     & \textbf{7.60} \\
ET  & \HL{HD95}  & 6.10            & 6.20  & \textbf{5.80}$^\dagger$ & 6.10  & 6.10     & 5.90 \\ \cline{1-8}
Avg.& \HL{HD95}  & 7.17            & 6.97  & 7.00$^\dagger$ & 7.43  & 6.77     & \textbf{6.63} \\ \hline
\end{tabular}
\end{adjustbox}
\end{table}

\begin{figure}
    \centering
    \includegraphics[width=\linewidth]{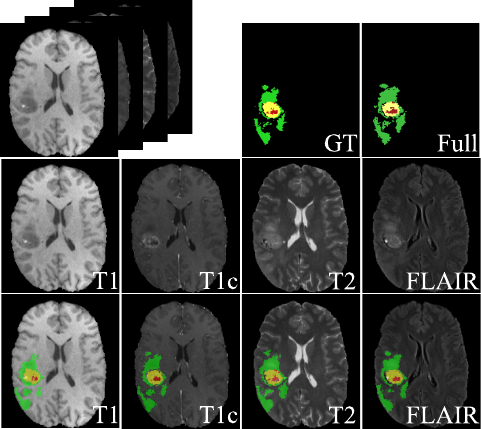}
    \caption{Visualization. Green: edema; yellow: enhancing tumor; and red: necrotic and non-enhancing tumor core. GT: Ground Truth. }
    \label{fig:vis_modal}
\end{figure}

Table~\ref{tab:2018} showcases the performance comparison between our method and four recent state-of-the-art methods for missing modality brain tumor segmentation on the BraTS 2018 dataset. In terms of overall average performance, our method achieves the best results across all three segmentation regions: 62.70\% for ET, 78.23\% for TC, and 86.47\% for WT, representing improvements of 1.93\%, 0.3\%, and 0.65\% over the second-best method, respectively. Figure~\ref{fig:vis_modal} shows example segmentation results by our method.

\begin{figure}
    \centering
    \includegraphics[width=1\linewidth]{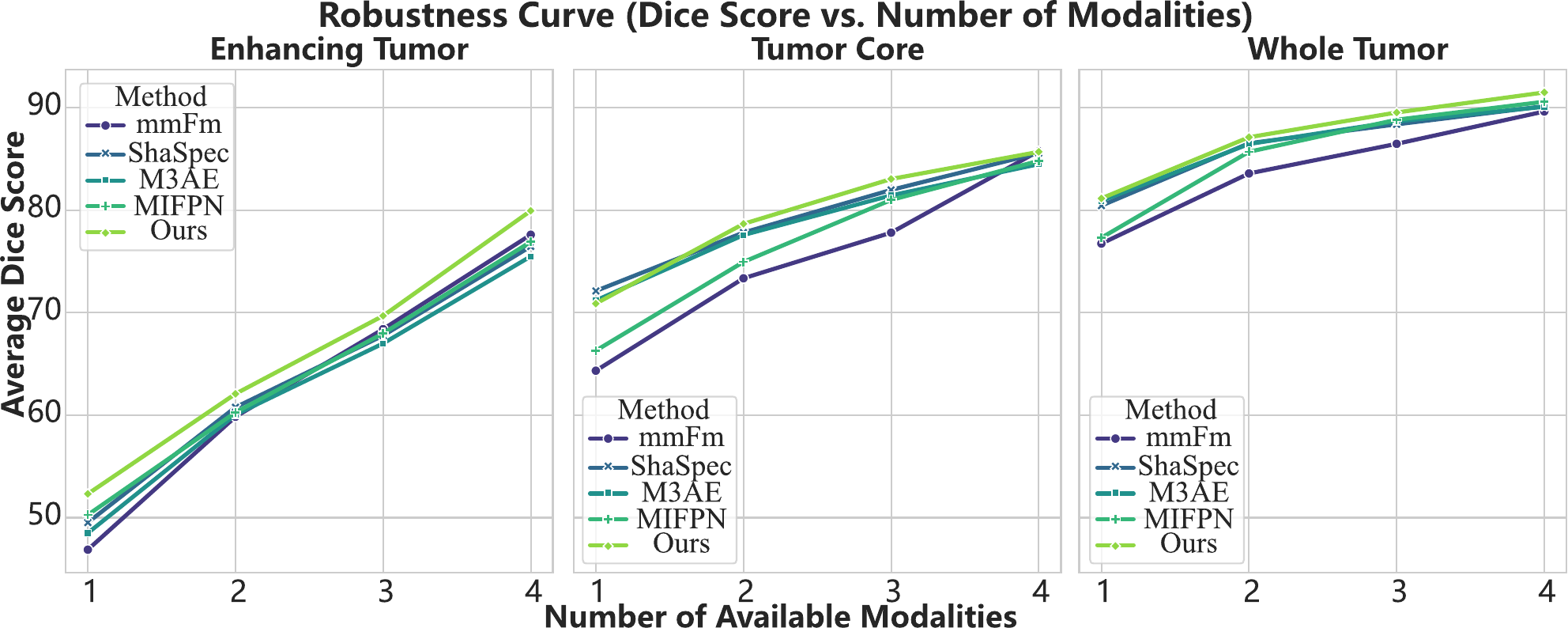}
    \caption{Comparison of average Dice scores across varying numbers of modalities.}
    \label{fig:curve}
\end{figure}

In single-modality scenarios, our method performs best in most cases, particularly achieving 90.40\% for WT segmentation with Modality 1, significantly outperforming other methods. Our method also demonstrates stable advantages in other modality combination scenarios. More intuitive results are presented in Figure~\ref{fig:curve}.

Table~\ref{tab:2020} presents the average performance comparison between our method and four competing methods on the BraTS 2020 dataset. The results indicate that our method achieves significant performance improvements across all segmentation regions: 87.21\% for WT, an improvement of 0.31\% over the best competitor M3AE; 82.54\% for TC, an improvement of 3.44\%; and 65.93\% for ET, an improvement of 4.23\%. The overall average performance reaches 78.56\%, which is a 2.66\% improvement over the second-best method M3AE. This significant enhancement validates the effectiveness and robustness of our proposed method.

\subsection{Ablation Study}

\begin{table}[]
\caption{Ablation Study. The experiments were conducted on BraTS 2018. K1 and K2 represent HMSD and DMCD, respectively. Metric: Dice Score.}
\label{tab:abla}
\centering
\begin{tabular}{cc|ccc|c}
\hline
K1           & K2           & Whole & Core  & Enhancing & Mean  \\ \hline
              & \ding{55}    & 84.68 & 76.79 & 61.34     & 74.27 \\
\ding{55}    &              & 85.26 & 77.48 & 62.36     & 75.03 \\
\ding{55}    & \ding{55}    & 85.03 & 74.61 & 61.48     & 73.71 \\ \hline
\checkmark   & \checkmark   & \textbf{86.44} & \textbf{78.23} & \textbf{62.70}     & \textbf{75.79} \\ \hline
\end{tabular}
\end{table}

To validate the effectiveness of each key component in our proposed framework, we conducted detailed ablation studies on the BraTS 2018 dataset. Table~\ref{tab:abla} shows the performance under different combinations of components, where K1 and K2 represent the HMSD and DMCD in our framework, respectively.

When the K1 component is removed, the average performance drops to 75.03\%, a decrease of 0.76\% compared to the full model, with a slight performance drop observed in ET segmentation. The K2 component contributes more noticeably to the ET segmentation task; its removal causes ET performance to drop from 62.70\% to 61.34\%, a decrease of 1.36 percentage points. When both K1 and K2 are removed, the model performance further drops to 73.71\%, indicating that both components positively contribute to the model's performance and exhibit a certain degree of synergistic effect.

\subsection{Feature Selection for Distillation}

To validate the effectiveness of feature selection within our proposed HMSD and DMCD mechanisms, and to investigate the core question of "which features are most suitable for knowledge transfer between modalities," we designed comprehensive ablation experiments. This section systematically compares the impact of different feature selection strategies on the model's final performance.

In our model design, specific features $f_{spec}^{m}$ and shared features $f_{shared}^{m}$ are extracted for each modality, which are then concatenated to form the fused modality feature $f_{fused}^{m}$. To explore which type of feature is more suitable for inter-modal knowledge transfer, we designed the following experiment: we implemented our two proposed distillation strategies (HMSD and DMCD) using $f_{spec}^{m}$, $f_{shared}^{m}$, and $f_{fused}^{m}$ separately. We report the experimental results for using different feature types with different distillation schemes.

As shown in Table~\ref{tab:feat}, the distillation mechanism can effectively address the missing modality problem, but the choice of features leads to differences in performance. The strategy using only specific features $f_{spec}^{m}$ for distillation yields the smallest performance gain. This aligns with our theoretical analysis: $f_{spec}^{m}$ is highly modality-specific. When a modality is missing, its corresponding specific features are absent, making knowledge transfer based on them unstable and poorly generalizable. The strategy using only shared features $f_{shared}^{m}$ performs better than using specific features. This suggests that the common, invariant representations learned from all modalities serve as a very robust carrier of knowledge, providing reliable and highly generalizable supervisory signals when modalities are missing. The fused features contain the richest information. The strategy using only $f_{fused}^{m}$ performs better than the previous two, as it incorporates both shared and specific information, providing the student model with soft targets closest to the final decision.

\begin{table}[]
\caption{Ablation study on the types of features used for distillation: Shared, Specific, and Fusion. Experiments are performed on the BraTS 2018 dataset. Dist.Type: Feature type used in distillation. Metric: Dice Score.}
\label{tab:feat}
\begin{adjustbox}{max width=\columnwidth}
\begin{tabular}{ccc|ccc|c}
\hline
Dist.Type                 & HMSD         & DMCD         & Whole & Core  & Enhancing & Mean                      \\ \hline
\multirow{2}{*}{Shared}   & $\checkmark$ &              & 82.71 & 75.28 & 59.89     & 72.63                     \\
                          &              & $\checkmark$ & 83.06 & 75.84 & 59.32     & 72.74                     \\
                          & $\checkmark$ & $\checkmark$ & 83.75 & 76.41 & 60.43     & 73.53                     \\ \hline

\multirow{2}{*}{Specific} & $\checkmark$ &              & 80.58 & 72.19 & 56.58     & 69.78                     \\
                          &              & $\checkmark$ & 81.57 & 72.80 & 58.13     & 70.83                     \\
                          & $\checkmark$ & $\checkmark$ & 80.17 & 71.04 & 56.50     & 69.24                     \\ \hline

\multirow{2}{*}{Fusion}   & $\checkmark$ &              & 84.68 & 76.79 & 61.34     & \multicolumn{1}{l}{74.27} \\
                          &              & $\checkmark$ & 85.26 & 77.48 & 62.36     & \multicolumn{1}{l}{75.03} \\
                          & $\checkmark$ & $\checkmark$ & \textbf{86.44} & \textbf{78.23} & \textbf{62.70}     & \textbf{75.79}                     \\ \hline
\end{tabular}
\end{adjustbox}
\end{table}

\subsection{Path Construction Strategy Analysis} 
\label{subsec:path}

A critical design choice within our DMCD framework lies in the strategy for constructing the decremental path, which sequence governing the order of modality removal. To rigorously validate the effectiveness of our proposed criticality-based removal strategy, whereby the most critical (highest-scoring) modality is removed at each step, we conducted a comprehensive ablation study comparing it against two alternative approaches. The first alternative, termed Min-Criticality, reverses our logic by prioritizing the removal of the least critical modality first, thereby presenting the model with a progressively more difficult but initially gentler learning trajectory. The second alternative, a Random-Path strategy, severs modality connections in a stochastic order. 

\begin{table*}[]
\caption{Results on BraTS 2018. Modalities 1-4 denote FLAIR, T1, T1c, and T2, respectively. Random and Min-C correspond to the Random-Path and Min-Criticality strategy, respectively. Metric: Dice score.}
\label{tab:path-full}
\centering
\begin{adjustbox}{max width=\textwidth}
\begin{tabular}{c|ccc|ccc|ccc}
\hline
\multirow{2}{*}{\textbf{Modalities}} & \multicolumn{3}{c|}{\textbf{Enhancing Tumor}}        & \multicolumn{3}{c|}{\textbf{Tumor Core}}             & \multicolumn{3}{c}{\textbf{Whole Tumor}}           \\ \cline{2-10} 
                                     & Random  & Min-C   & Default       & Random  & Min-C & Default     & Random  & Min-C & Default   \\ \hline
1                                    & 44.18   & 43.35     & \textbf{45.48}         & 70.33   & 69.62 & \textbf{70.89}         & 89.64   & 88.71 & \textbf{90.40}       \\
2                                    & 42.94   & 41.67     & \textbf{42.82}         & 64.62   & 63.35 & \textbf{65.20}         & 73.95   & 73.46 & \textbf{74.17}       \\
3                                    & 73.23   & 73.42     & \textbf{74.22}         & 77.30   & 77.06 & \textbf{78.05}         & 75.25   & 74.84 & \textbf{75.52}       \\
12                                   & 45.29   & 43.39     & \textbf{45.58}         & \textbf{73.83}   & 72.28 & 73.60         & 88.39   & 88.71 & \textbf{88.93}       \\
13                                   & 76.17   & 75.49     & \textbf{76.84}         & 83.21   & 83.52 & \textbf{84.31}         & 89.77   & 89.50 & \textbf{90.46}       \\
23                                   & 76.98   & 76.48     & \textbf{77.01}         & 83.33   & 83.03 & \textbf{84.02}         & 78.48   & 78.53 & \textbf{78.63}       \\
123                                  & 77.10   & 77.53     & \textbf{77.63}         & 84.75   & 84.95 & \textbf{85.28}         & 88.93   & 88.42 & \textbf{89.20}       \\
Avg.                                 & 62.27   & 61.62     & \textbf{62.80}         & 76.77   & 76.26 & \textbf{77.34}         & 83.49   & 83.17 & \textbf{83.90}       \\ \hline
\end{tabular}
\end{adjustbox}
\end{table*}

We implement all three path construction strategies within the same DMCD framework, keeping all other components identical. The experiment is conducted on the BraTS 2018 dataset. For each strategy, we start with the full modality combination, and then construct a decremental path by iteratively removing one modality according to the respective strategy. Finally, we perform sequential knowledge distillation along the constructed path and evaluate the final model's performance on the test set. For concise, we selected the Flair, T1, and T1c combination for presentation in the Table~\ref{tab:path-full}. 

Table~\ref{tab:path-full} presents the comparative performance of the three path construction strategies under different evaluation conditions. The proposed DMCD (argmax) achieves the best performance across all evaluation scenarios, with an average Dice of 74.68 compared to 74.17 for Random-Path and 73.68 for Min-Criticality. 

These results collectively validate our hypothesis that deliberately constructing paths by removing the most critical modalities first creates more effective training scenarios for knowledge distillation. The DMCD approach forces the model to learn how to compensate for the loss of irreplaceable information, resulting in better generalization when faced with real-world modality missing scenarios. In contrast, Min-Criticality focuses on easier cases where redundant information is lost, providing less valuable learning signals, while Random-Path lacks the systematic approach needed to build robust multimodal representations.

This experiment provides strong evidence that the specific ordering of modality removal is crucial to the success of decremental distillation, and our criticality-based strategy offers a principled approach to constructing optimal decremental paths for multimodal learning.

\subsection{Task-aware Criticality Estimation}
\label{subsec:taskaware}

\added{As discussed in Section~\ref{sec:DMCD}, the default DMCD adopts a cosine-similarity-based 
criticality score to estimate modality replaceability. 
To further investigate whether explicitly task-aware signals can improve 
the decremental path construction, we implement a gradient-based criticality variant.}

\paragraph{\added{Gradient-based criticality}}
\added{For each modality $m_j$, we define the task-aware score as:}
\begin{equation}
s_{\text{grad}}(m_j) =
\left\| \frac{\partial \mathcal{L}}{\partial f^{m_j}_{\mathrm{fused}}} \right\|_2,
\end{equation}
\added{where $\mathcal{L}$ denotes the segmentation loss and 
$f^{m_j}_{\mathrm{fused}}$ represents the fused feature of modality $m_j$.
This score reflects the instantaneous sensitivity of the task objective to each modality.
Importantly, it can be computed using a single backward pass of the original loss, 
without introducing additional forward evaluations or auxiliary objectives.}

\added{We use this score to construct decremental paths in DMCD 
by iteratively removing the modality with the highest gradient magnitude.
All other components remain unchanged to ensure a fair comparison.}

\begin{table}[t]
\caption{
  \added{
    Comparison of different criticality estimation strategies within DMCD on BraTS 2018. Metric: Dice (\%).
  }
}
\label{tab:crit}
\centering
\begin{tabular}{l|ccc|c}
\hline
Criticality Type & WT & TC & ET & Mean \\
\hline
Random              & 85.03          & 74.61          & 61.48          & 73.71 \\
Gradient-based      & 85.12          & 77.95          & 62.41          & 75.16 \\
Cosine-based (Ours) & \textbf{86.47} & \textbf{78.23} & \textbf{62.70} & \textbf{75.79} \\
\hline
\end{tabular}
\end{table}

\added{Table~\ref{tab:crit} presents the results.
The gradient-based strategy achieves competitive performance 
compared to the redundancy-based cosine criterion, 
demonstrating that DMCD is not restricted to a specific criticality formulation.
However, the cosine-based approach consistently yields slightly higher 
average Dice scores across all three regions.}

\added{These results indicate that task-aware signals can be naturally integrated 
into DMCD, while redundancy-based criticality offers 
a favorable trade-off between stability and robustness 
for unified multimodal training.}

\subsection{Qualitative Results}

\begin{figure}
    \centering
    \includegraphics[width=1\linewidth]{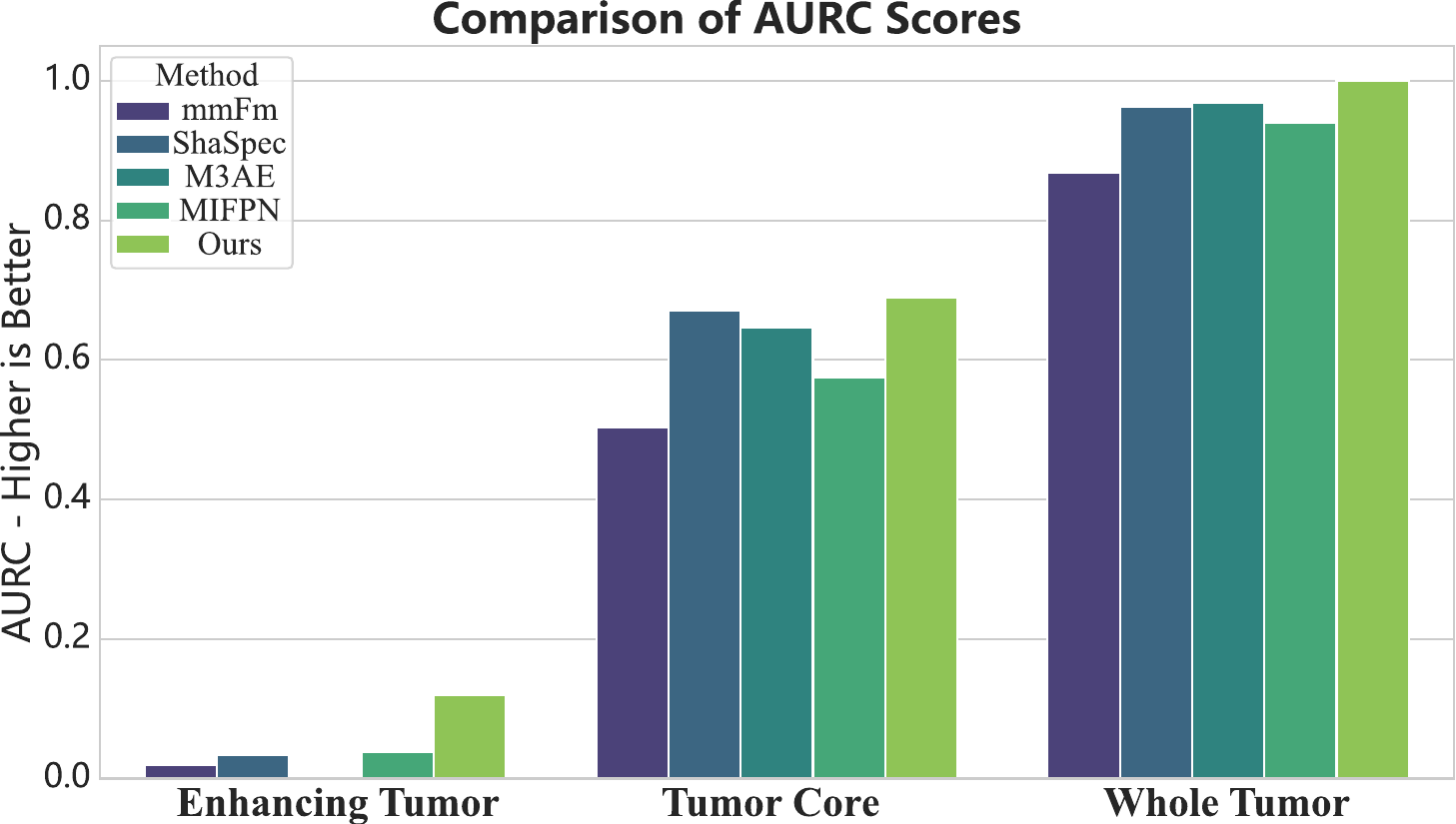}
    \caption{Comparison of AURC scores among our method and four baseline methods on the three segmentation tasks: ET, TC, and WT. AURC evaluates the overall performance and robustness of a model under varying degrees of modality missing by computing the area under the curve of average Dice scores with respect to the number of available modalities. A higher AURC indicates better stability and performance, especially when modalities are missing.}
    \label{fig:aurc}
\end{figure}

\deleted{To more comprehensively evaluate the overall performance and stability of the model under varying degrees of modality missing scenarios, we introduce the Area Under the Robustness Curve (AURC) as a key quantitative evaluation metric. While the traditional Dice score can show the model's performance under specific modality combinations, it is difficult to measure its overall robustness when facing gradually reducing information.}

\deleted{The calculation process of the AURC metric is as follows. First, we group all 15 modality combinations based on the number of available modalities. Second, we calculate the average Dice score for all test cases within each group, resulting in four key performance points. Finally, we plot the "Robustness Curve" with the number of available modalities on the x-axis and the corresponding average Dice score on the y-axis. The AURC is the area under this curve. A higher AURC value indicates that the model's robustness curve is positioned higher overall, signifying that the model maintains superior comprehensive performance both when information is sufficient and when it is severely lacking.}

\deleted{Figure~\ref{fig:aurc} shows the comparison of AURC scores between our method and the other four methods on the three segmentation tasks: ET, TC, and WT. It can be clearly observed from the figure that our proposed method achieves the highest AURC values across all three tasks. It not only performs excellently when all four modalities are available but, more importantly, its performance declines more gradually when facing the challenge of one or more missing modalities, consistently maintaining a performance advantage over other methods. The consistently leading AURC values fully validate the effectiveness and stability of our framework in handling incomplete multi-modal data, which is of great significance for practical clinical applications.}

\added{To more comprehensively evaluate the overall performance and stability of the model under varying degrees of modality missing scenarios, we introduce the Area Under the Robustness Curve (AURC) as a key quantitative evaluation metric. While the traditional Dice score can show the model's performance under specific modality combinations, it is difficult to measure its overall robustness when facing gradually reducing information.}

\added{Definition and Calculation. Following the notation in Section X, let $\mathcal{S}_N^k$ denote the set of modality combinations containing exactly $k$ available modalities, where $k \in \{1, 2, 3, ..., N\}$. In our four-modality setting ($N=4$), we have $|\mathcal{S}_4^1|=4$, $|\mathcal{S}_4^2|=6$, $|\mathcal{S}_4^3|=4$, and $|\mathcal{S}_4^4|=1$, yielding 15 combinations in total. The group-wise average Dice score is computed as:}
$$\bar{D}_k = \frac{1}{|\mathcal{S}_N^k|}\sum_{c \in \mathcal{S}_N^k} \text{Dice}(c)$$

\added{The "Robustness Curve" is formed by plotting the four points $(k, \bar{D}_k)$, with the number of available modalities on the x-axis and the corresponding average Dice score on the y-axis. The AURC is calculated using the trapezoidal rule:}
$$\text{AURC} = \frac{1}{2}\bar{D}_1 + \bar{D}_2 + \bar{D}_3 + \frac{1}{2}\bar{D}_4$$

\added{A higher AURC value indicates that the model's robustness curve is positioned higher overall, signifying that the model maintains superior comprehensive performance both 
when information is sufficient and when it is severely lacking. Notably, this formulation assigns higher weights to intermediate modality availability levels (2 and 3 modalities), 
which aligns with common clinical scenarios where partial information is often available.}

\added{Figure~\ref{fig:aurc} shows the comparison of AURC scores between our method and the other four methods on the three segmentation tasks: ET, TC, and WT. 
Our method achieves the highest AURC values across all three regions. % : 197.92 (ET), 240.02 (TC), and 263.00 (WT)
% Compared to the second-best methods, CCSD improves AURC by +6.06 on ET (vs. MIFPN), +1.36 on TC (vs. ShaSpec), and +2.35 on WT (vs. M3AE). 
The most substantial improvement is observed on the challenging Enhancing Tumor region, where accurate segmentation heavily relies on complementary information from multiple modalities. 
As shown in the figure, while all methods experience performance degradation as modalities are removed, our method exhibits a flatter decline, consistently maintaining a performance advantage. 
These results validate the effectiveness and stability of our framework in handling incomplete multi-modal data, which is of great significance for practical clinical applications.}

\subsection{\added{Limitations and Future Work}}

\added{\textbf{Modality interaction modeling.}
While CCSD achieves robust performance under arbitrary missing-modality settings, cross-modality interactions are modeled in a relatively constrained manner. Interactions are primarily realized through feature concatenation and lightweight composition, and the DMCD criticality score relies on pairwise cosine similarity. Consequently, higher-order or non-linear dependencies among multiple modalities are not explicitly modeled, which may limit expressiveness in cases with highly asymmetric or uniquely informative interactions.}

\added{\textbf{Missing-modality assumptions and robustness.}
Absent modalities are represented solely via shared features to maintain architectural consistency between training and inference. This design assumes that shared representations can sufficiently compensate for missing cross-modality cues, which may be less effective in extreme cases. In addition, we do not explicitly evaluate joint scenarios involving severe signal-level corruption and modality absence.}

\added{Future work will explore more expressive yet stable interaction mechanisms, task-aware modality criticality measures, unified modeling of modality absence and corruption, and broader cross-site validation.}

\section{Conclusion}

In this paper, we tackle the challenge of handling missing modalities in multi-modal brain tumor segmentation by building upon the shared-specific encoder-decoder architecture, which effectively disentangles modality-invariant and modality-specific features. Our primary contribution lies in enhancing this framework with two novel self-distillation strategies: HMSD, which facilitates knowledge transfer from full modality sets to subsets across hierarchy levels, and DMCD, which simulates progressive modality failure scenarios to boost robustness against information loss. This integrated approach elegantly handles arbitrary input modality combinations without requiring architectural changes during inference.

Our work presents a knowledge-oriented solution to the missing modality problem, with significant potential for reliable application in clinical settings where incomplete data is prevalent. For future research, we plan to explore advanced feature fusion mechanisms and extend our proposed distillation paradigm to other multi-modal medical image analysis tasks, further enhancing the utility of this approach in diverse clinical contexts.

\appendix

\section{Feature Dimensions}

Let the input tensor be denoted as  
\[
X \in \mathbb{R}^{N \times D \times W \times H},
\]
where \(N\) is the number of modalities, and \(D\), \(W\), and \(H\) represent the spatial dimensions of the volume.

The shared encoder produces modality-wise shared representations:
\[
f^{m}_{\mathrm{shared}} \in 
\mathbb{R}^{C_s \times D' \times W' \times H'},
\]
where \(C_s\) denotes the number of shared feature channels, and \(D'\), \(W'\), and \(H'\) denote the spatial dimensions of the encoder outputs.

Similarly, each modality-specific encoder generates:
\[
f^{m}_{\mathrm{spec}} \in 
\mathbb{R}^{C_s \times D' \times W' \times H'}.
\]

After feature fusion, we obtain:
\[
f^{m}_{\mathrm{fused}} \in 
\mathbb{R}^{C_s \times D' \times W' \times H'}.
\]

The fused features from all modalities are then concatenated along the channel dimension, yielding:
\[
Z_{\mathcal{S}_i} \in 
\mathbb{R}^{(N \cdot C_s) \times D' \times W' \times H'}.
\]

Finally, the decoder maps \(Z_{\mathcal{S}_i}\) to segmentation logits:
\[
Y \in \mathbb{R}^{K \times D \times W \times H},
\]
where \(K\) denotes the number of target classes.

\bibliographystyle{elsarticle-num-names} 
\bibliography{refs}

\end{document}